\documentclass[sn-mathphys-num]{sn-jnl}

\usepackage{graphicx}%
\usepackage{multirow}%
\usepackage{amsmath,amssymb,amsfonts}%
\usepackage{amsthm}%
\usepackage{mathrsfs}%
\usepackage[title]{appendix}%
\usepackage{xcolor}%
\usepackage{textcomp}%
\usepackage{manyfoot}%
\usepackage{booktabs}%
\usepackage{algorithm}%
\usepackage{algorithmicx}%
\usepackage{algpseudocode}%
\usepackage{listings}%
\usepackage{natbib}
\usepackage{siunitx}

\begin{document}

\title[Carotid Artery Plaque Analysis in 3D Based on Distance Encoding in Mesh Representations]{Carotid Artery Plaque Analysis in 3D Based on Distance Encoding in Mesh Representations}


\author*[1]{\fnm{Hinrich} \sur{Rahlfs}}\email{hinrich.rahlfs@dhzc-charite.de}

\author[1,2,3]{\fnm{Markus} \sur{Hüllebrand}}

\author[4]{\fnm{Sebastian} \sur{Schmitter}}

\author[5]{\fnm{Christoph} \sur{Strecker}}

\author[5]{\fnm{Andreas} \sur{Harloff}}

\author[1,2,3]{\fnm{Anja} \sur{Hennemuth}}

\affil[1]{\orgdiv{Institute of Computer-Assisted Cardiovascular Medicine}, \orgname{Deutsches Herzzentrum der Charité}, \orgaddress{\city{Berlin}, \country{Germany}}}

\affil[2]{\orgname{Fraunhofer MEVIS}, \orgaddress{\city{Bremen, Germany}, \country{Germany}}}

\affil[3]{\orgname{DZHK (German Centre for Cardiovascular Research)}, \orgaddress{\city{Partner Site Berlin}, \country{Germany}}}

\affil[4]{\orgname{Physikalisch-Technische Bundesanstalt (PTB)}, \orgaddress{\city{Berlin}, \country{Germany}}}

\affil[5]{\orgdiv{Department of Neurology and Neurophysiology, Faculty of Medicine}, \orgname{Medical Center—University of Freiburg}, \orgaddress{\city{Freiburg}, \country{Germany}}}

\abstract{\textbf{Purpose:} Enabling a comprehensive and robust assessment of carotid artery plaques in 3D through extraction and visualization of quantitative plaque parameters. These parameters have potential applications in stroke risk analysis, evaluation of therapy effectiveness, and plaque progression prediction.

\textbf{Methods:}
We propose a novel method for extracting a plaque mesh from 3D vessel wall segmentation using distance encoding on the inner and outer wall mesh for precise plaque structure analysis. A case-specific threshold, derived from the normal vessel wall thickness, was applied to extract plaques from a dataset of 202 T1-weighted black-blood MRI scans of subjects with up to 50\% stenosis. Applied to baseline and one-year follow-up data, the method supports detailed plaque morphology analysis over time, including plaque volume quantification, aided by improved visualization via mesh unfolding.

\textbf{Results:} We successfully extracted plaque meshes from 341 carotid arteries, capturing a wide range of plaque shapes with volumes ranging from 2.69$\mu\text{l}$ to 847.7$\mu\text{l}$. The use of a case-specific threshold effectively eliminated false positives in young, healthy subjects. 

\textbf{Conclusion:} The proposed method enables precise extraction of plaque meshes from 3D vessel wall segmentation masks enabling a correspondence between baseline and one-year follow-up examinations. Unfolding the plaque meshes enhances visualization, while the mesh-based analysis allows quantification of plaque parameters independent of voxel resolution.
}


\maketitle
\begin{figure}[H]
  \centering
  \includegraphics[width=0.9\linewidth]{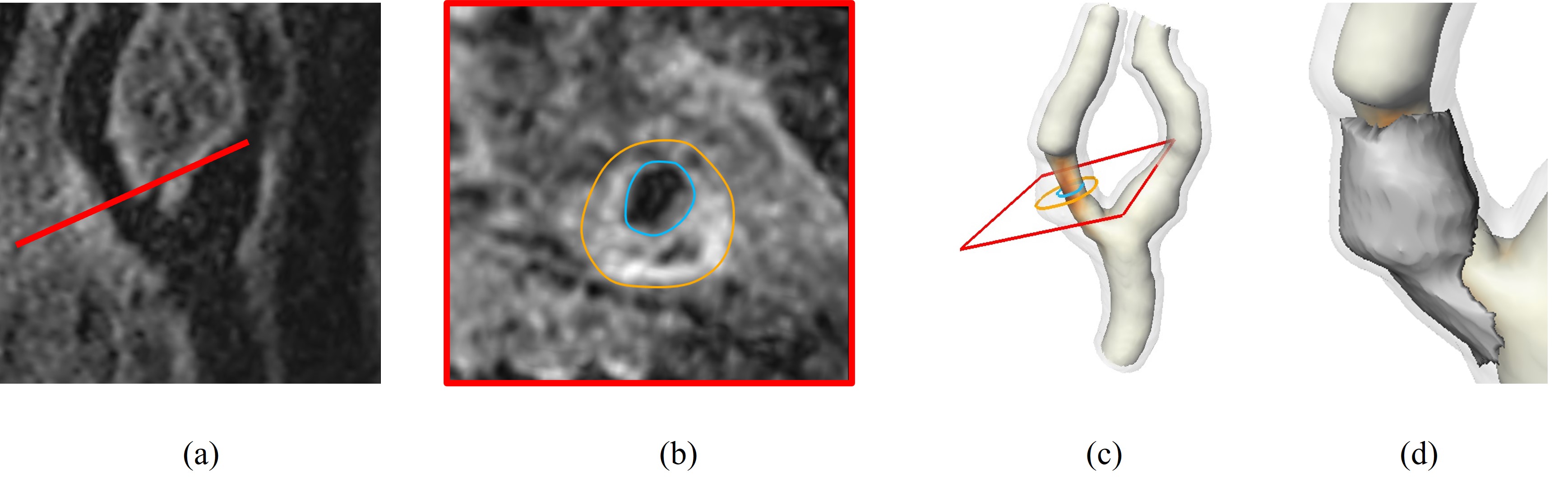}
  \caption{(a) View of an atherosclerotic carotid artery in T1-weighted 3D BB-MRI with cross-section MPR. (b) Segmented 2D cross-section of the carotid artery which can be used for the computation of wall area and maximum wall thickness. (c) 3D segmentation of the carotid wall with the VWT color-coded on the lumen surface. (d) Plaque extracted from the 3D segmentation which can be used for the quantification of plaque volume, plaque surface and compactness.}
  \label{fig:introduction}
\end{figure}

\section{Introduction}\label{sec1}

Atherosclerosis of the carotid artery and the resulting internal carotid artery stenosis are a major risk factor for stroke \cite{nagai2001significance}, which is a leading cause of disability and mortality worldwide \cite{campbell2019ischaemic}. Magnetic resonance imaging (MRI) can be used for evaluating the effectiveness of therapy, assessing risk, and predicting plaque progression in the carotid artery \cite{lee2008early, saam2007predictors, bijari2014carotid, strecker2020carotid, strecker2021carotid, saba2021review, van2022carotid, brunner2021associations}. Earlier studies \cite{lee2008early, saam2007predictors, bijari2014carotid, strecker2020carotid, strecker2021carotid} relied on manually segmented 2D cross-sectional images (Fig.~\ref{fig:introduction}.b), which have significant limitations. These methods can only assess parameters such as vessel wall thickness (VWT) and vessel wall area. Additionally, the placement of the cross-section influences these measurements, introducing variability and limiting the comprehensiveness of the analysis. Fig.~\ref{fig:introduction}.a shows the placement of one cross-section close to the maximum stenosis. The exact position of maximal stenosis cannot be determined on this 2D image.

3D segmentation of the carotid artery is possible, using CNN-based approaches trained on pseudo-labels. They can be generated through interpolation between 2D segmentations \cite{hu2022label} or by using adversarial 2D networks \cite{rahlfs2024sparse, li2024carotid}. A further refinement using a 2D CNN can be beneficial\cite{lavrova2023ur}.
This facilitates the automatic extraction of 3D plaque parameters, such as plaque surface area, plaque volume, and compactness. Especially the plaque volume has the potential to be a better parameter for stroke risk assessment than VWT, if the automatic quantification is implemented \cite{saba2021review, van2022carotid}. It further opens new possibilities for improved visualization of the distribution of vessel wall features in 3D (Fig.~\ref{fig:introduction}.c). While the measurement of these parameters in 3D is straight forward, visualization is challenging, since not all parts of the wall/plaque surface can be viewed at the same time. 

We address these challenges by extracting a plaque mesh from the 3D vessel wall segmentation. This mesh representation can be used for the quantification of 3D plaque parameters such as the volume of plaque constituents. We also show a mesh-based unfolding of the plaque surface for the visualization of the spatial plaque parameter distribution.

\section{Methods}
\subsection{Data} For method development and evaluation we used a total of 212 T1-weighted black-blood (BB)-MRI scans of the neck region, covering the left and right carotid arteries close to the bifurcation. The scans were obtained using a Siemens Prisma 3T scanner with a T1-weighted 3D Turbo Spin Echo sequence (3D-SPACE) that incorporates fat saturation and dark blood preparation. Detailed descriptions of the data are provided by Strecker et al. \cite{strecker2020carotid,strecker2021carotid}. The scans were divided into a stenosis-set and a healthy-set.

The stenosis-set consist of 202 BB-MRI scans. 121 subjects were scanned in a baseline examination and 81 of these subjects were scanned in a one year follow-up examination. All subjects of the stenosis-set were diagnosed with hypertension, each presenting at least one additional cardiovascular risk factor and a plaque measuring $\geq$~1.5 mm in internal carotid artery (ICA) and/or common carotid artery (CCA) in ultrasound. All subjects have an ICA stenosis smaller than 50\%, as defined by NASCET criteria \cite{von2012grading}. The average age of subjects in the stenosis-set was 70.7 years.

The healthy-set consists of 10 BB-MRI scans of healthy subjects with an average age of 34.1 years. There was no follow-up examination performed for healthy subjects.

\subsection{Plaque Mesh Extraction}

\begin{figure}[h]
\centering
\includegraphics[width=\textwidth]{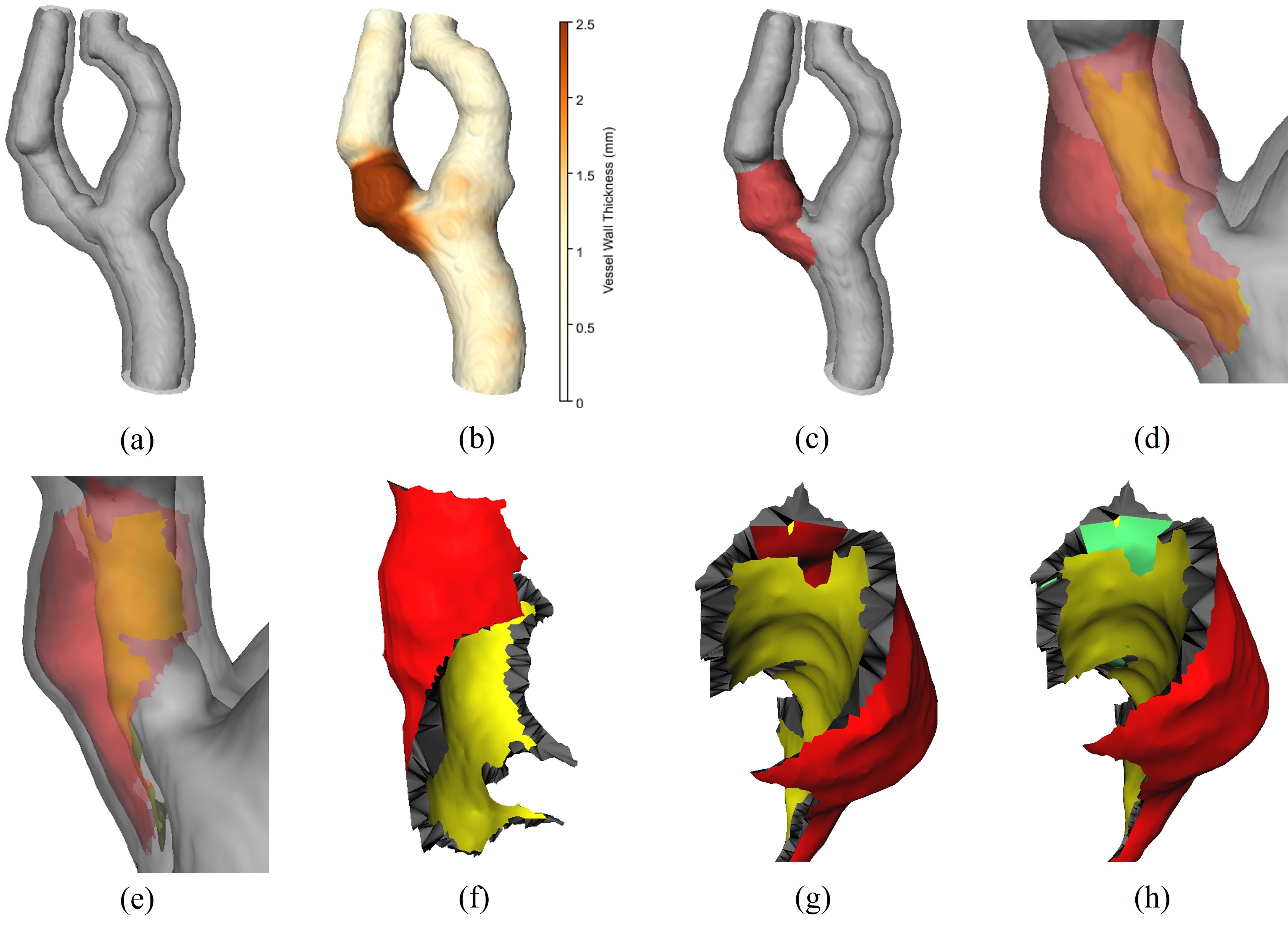}
\caption{Process of plaque mesh extraction. (a) Inner and outer wall mesh created by the marching cubes algorithm \cite{lewiner2003efficient} and smoothed with Laplacian smoothing. (b) Outer wall mesh color-coded with the distance to the inner wall mesh. (c) Plaque region (red) detected on the outer wall mesh. (d) Corresponding plaque region on the inner wall mesh (yellow). (e) Inner and outer plaque border meshes, created by shifting all vertices of the plaque regions by half the mean VWT of the non-plaque vertices. (f) Boundary mesh created between inner and outer plaque border meshes. (g) The created boundary is incomplete due to two submeshes in the inner plaque border mesh (yellow). (h) Fixed mesh (green) using MeshFix \cite{attene2010lightweight}.}\label{fig:plaque_extraction}
\end{figure}

\paragraph{3D Artery Mesh Creation}
The carotid artery wall and lumen were segmented using a 3D U-Net. It was trained on 3D pseudo-labels that were created using an auxiliary 2D U-Net as described by Rahlfs et al. \cite{rahlfs2024sparse}. The resulting voxel mask was transformed into triangle meshes using the marching cubes algorithm \cite{lewiner2003efficient} on the lumen mask for the inner wall mesh and on the union of the lumen and wall mask to obtain the outer wall mesh.  Laplacian smoothing was applied with 10 iterations and a relaxation factor of 0.2. The result is shown in Fig.~\ref{fig:plaque_extraction}.b.

\paragraph{Plaque Region Extraction}
The plaque region was extracted using the distances between the vertices $X_{outer}$ of the outer wall and the inner wall mesh $M_{inner}$ (Fig.~\ref{fig:plaque_extraction}.b). The plaque vertices $X_{plaque}$ were detected by Eq.\ref{eq:plaque_vertices} via the plaque threshold $pt$. We defined the relevant plaque region as the biggest submesh with all face vertices in $X_{plaque}$. Submeshes with an area smaller than 10 $mm^2$ were discarded. An example result is shown in Figure~\ref{fig:plaque_extraction}.c.
\begin{equation}
X_{plaque} = \{ x \in X_{outer} \mid \text{dist}(x, M_\text{inner}) > pt \}
\label{eq:plaque_vertices}
\end{equation}

We evaluated two approaches for the threshold $pt$. A global threshold $pt=1.496~mm$ was chosen based on Zhang et al. \cite{zhang2018carotid}, who state that the mean VWT of the CCA measured with BB-MRI is 0.98~mm and the standard deviation is 0.2~mm. Leading to 1.496~mm as the border of the 2.58$\sigma$ environment.

The second approach was the use of a case-specific threshold. This threshold was computed iteratively using Eq.~\ref{eq:iterative_threshold}~and~\ref{eq:iterative_threshold_exp1}. As initialization, all outer vertices are normal vertices (Eq.~\ref{eq:iterative_threshold_exp2}). The threshold is computed by Eq.\ref{eq:iterative_threshold_exp1}, and $X_{normal}$ is updated using the new threshold. The two steps are repeated until convergence. We evaluated this approach on 12 carotid arteries for $k\in\{2, 2.5, 3, 3.5, 4, 4.5\}$ and chose a k that was able to extract all relevant plaques without overestimating the plaque region.

\begin{equation}
X_{normal}^{t+1} = \{ x \in X_{outer} \mid \text{dist}(x, M_\text{inner}) \leq pt(X_{normal}^{t}) \}
\label{eq:iterative_threshold}
\end{equation}
where, 
\begin{equation}
pt(X_{normal}^{t}) = \mu(X_{normal}^{t}) + k*\sigma(X_{normal}^{t}) \label{eq:iterative_threshold_exp1}
\end{equation}
\begin{equation}
X_{normal}^{0} = X_{outer} \label{eq:iterative_threshold_exp2}
\end{equation} 

The plaque region was projected on the inner wall mesh by choosing the submesh where all vertices have the outer wall plaque region as the nearest neighbor. The result can be seen in Fig.~\ref{fig:plaque_extraction}.d.

\paragraph{Plaque Mesh Creation}
To include only the volume that was added due to increased VWT, the inner and outer plaque region vertices were shifted towards each other by half the mean VWT. This can either be 0.98~mm \cite{zhang2018carotid} if using the fixed threshold approach or the mean distances between $X_{normal}$ and $M_{outer}$ if using the case-specific approach. This leads to the inner and outer wall of the plaque as shown in Fig.~\ref{fig:plaque_extraction}.e.

As a next step, the borders of the inner and outer plaque meshes were stitched together as shown in Fig.~\ref{fig:plaque_extraction}.f. If one of the plaque meshes has more than one closed edge, this stitching does not work perfectly. Fig.~\ref{fig:plaque_extraction}.g shows an example, where there is a big inner plaque border (yellow) and small disconnected submesh in yellow at the top.

To ensure that all plaque meshes are watertight, we used the MeshFix algorithm of Attene et al. \cite{attene2010lightweight}. The result can be seen in Fig.~\ref{fig:plaque_extraction}.h with the green parts of the mesh added by MeshFix.

\subsection{Vessel Wall Thickness Visualization in 2D} Visualization of parameters on surfaces in 3D is not always intuitive and surface unfolding can improve the parameter visualization \cite{kreiser2018survey, eulzer2021automatic} To improve the visualization, we unfold the plaque region extracted from the outer vessel wall and display the VWT on the unfolded 2D meshes. To unfold the meshes, we use the LSCM algorithm \cite{levy2023least}.

\subsection{Geometric Plaque Parameters}
Following parameters are derived from the extracted plaque meshes:

\begin{itemize}
\item Plaque Volume ($V$): The volume enclosed by the plaque mesh

\item Diameter of Ideal Sphere ($D_{Sphere})$: The diameter of an ideal sphere with the same volume as the plaque. Calculated by Eq.~\ref{eq:sphere}

\item Plaque Surface Area ($A$): The area of the plaque mesh surface

\item Diameter of Ideal Circle ($D_{Circle})$: The diameter of an ideal circle with the same surface area as the plaque. Calculated by Eq.~\ref{eq:circle}

\item Compactness ($C$): Compactness of the plaque normalized to an ideal sphere as calculated by Eq.~\ref{eq:compactness}

\item Maximum Extent: The maximum distance between two vertices of the mesh

\begin{equation}
D_{Sphere} = \sqrt[3]{\frac{6*V}{\pi}}
\label{eq:sphere}
\end{equation}

\begin{equation}
D_{Circle} = 2*\sqrt[2]{\frac{A}{\pi}}
\label{eq:circle}
\end{equation}

\begin{equation}
C = \frac{36 \pi V^2}{A^3}
\label{eq:compactness}
\end{equation}

\end{itemize}

\subsection{Intensity Distribution Within Plaque Mesh}
To analyze the intensity distributions within plaques, all voxels whose center points fall inside the plaque mesh were identified. The T1-weighted BB-MRI intensities corresponding to these voxels were extracted and used to construct a histogram.

\subsection{Choice of Threshold Method}

We chose the case-specific threshold heuristically based on the evaluation of 6 MRI-scans. The plaque region was overestimated for $k=2$, we extracted similar plaque meshes for $k\in\{2.5,3,3.5\}$. For $k\in\{4, 4.5\}$, no plaque was extracted for arteries with stronger stenosis. Thus, we set the threshold to $k=3$.

To validate the plaque extraction and evaluate differences between the global and case-specific threshold, we applied the plaque extraction on the stenosis-set and on the healthy-set. If a plaque was extracted on one of the healthy-set arteries we considered this to be a false positive. A carotid artery of the stenosis-set where no plaque is extracted does not qualify as a false negative, as it is only known that one of the carotid arteries contains a plaque $\geq$ 1.5~mm. Therefore, we chose the method with less false positives for further experiments.

\section{Results}
Vessel wall segmentation, plaque extraction and plaque unfolding were applied to all BB-MRI datasets. Plaque meshes were extracted in 341 of 404 cases using the case-specific threshold and in 338 of 404 cases for the global threshold. In the datasets of healthy subjects the global threshold extracted a plaque in eight of 20 cases, where the ultrasound examination had not detected plaques. The application of the case-specific threshold resulted in no plaque in any of the healthy carotid arteries. 
Given that the case-specific threshold did not show any false positives, we used the case-specific threshold as a basis for ensuing analysis steps.

\begin{figure}
  \centering
  \includegraphics[width=\linewidth]{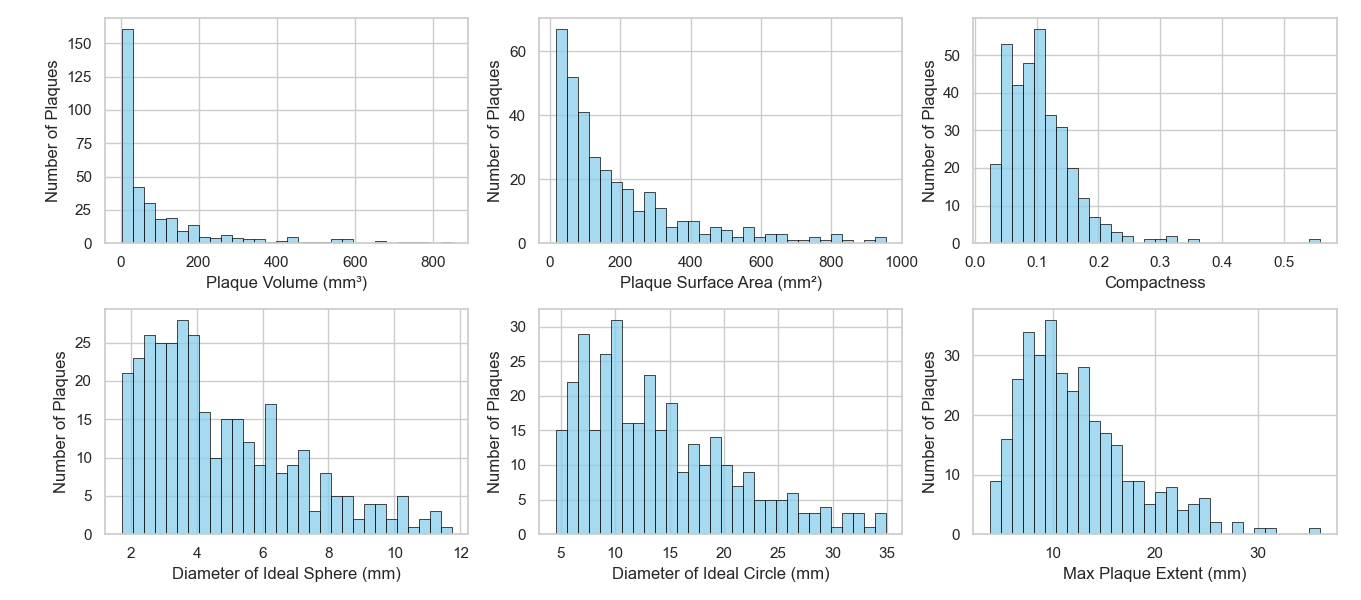}
  \caption{Distribution of plaque parameters. For the calculation of the parameters the 341 plaque meshes extracted with the case-specific threshold.}
  \label{fig:plaque_parameter_distribution}
\end{figure}
\subsection{Distribution of Geometric Plaque Parameters}
Fig.~\ref{fig:plaque_parameter_distribution} shows the quantitative geometric parameters calculated using the plaque meshes. 161 plaques have a plaque volume of less than 30.9~$mm^3$ and the maximum plaque volume is 847.7~$mm^3$. The diameter of the ideal sphere shows a much wider distribution than the plaque volume.
The average compactness of the plaques is 0.105 and the plaques can have an extent of up to 36.1~$mm$.

\begin{figure}
  \centering
  \includegraphics[width=\linewidth]{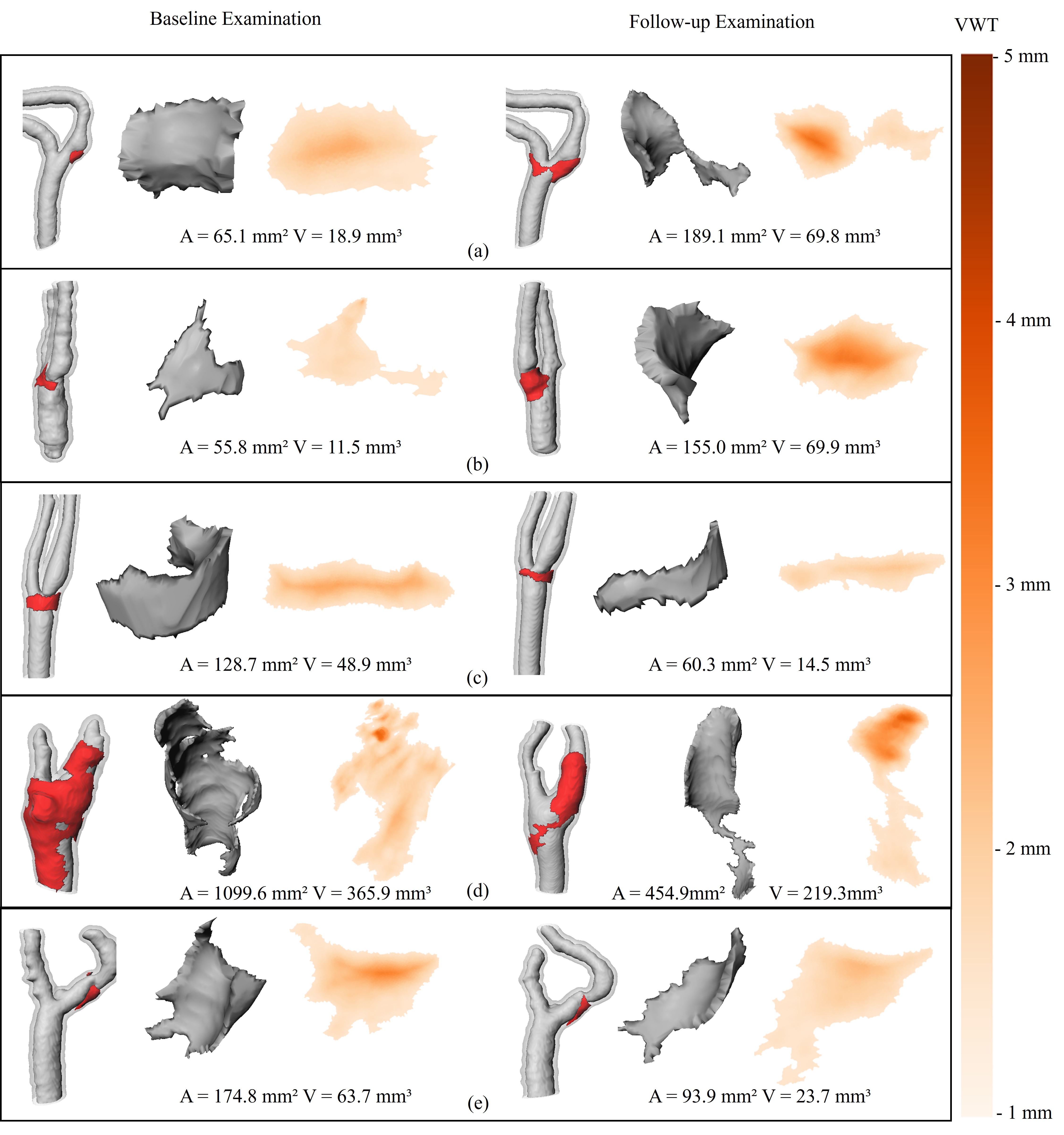}
  \caption{Plaque extraction using case-specific thresholds on the baseline and follow-up scans of five carotid arteries. For each examination, the inner and outer wall meshes are shown on the left with the plaque mesh added in red. In the middle, a different angle of the plaque mesh is shown. On the right, the unfolded outer plaque region mesh is shown color-coded with the vessel wall thickness (VWT). This allows a fast overview of the plaque development.}
  \label{fig:plaque-visualization}
\end{figure}

\subsection{Results of Baseline and Follow-up Examinations}
We chose five subjects with follow-up measurements for the qualitative evaluation and assessed plausibility and interpretability of the quantitative and visual results.  Fig.~\ref{fig:plaque-visualization} shows the inner wall mesh with the plaque mesh, the plaque mesh from a different angle, and the unfolded outer plaque region color-coded with the VWT. 

Fig.~\ref{fig:plaque-visualization}.a shows the results for a vessel with high tortuosity. The surface area ($A$), the VWT, and plaque volume (V) are increased in the follow-up examination. 

The second carotid artery (Fig.~\ref{fig:plaque-visualization}.b) has an irregular, wavelike surface. This case also shows a strong growth in plaque area, volume and VWT between baseline and follow-up.

In (Fig.~\ref{fig:plaque-visualization}.c), the extracted plaque in the follow-up examination is smaller. The plaque is extracted at the same position in both examinations.

Fig.~\ref{fig:plaque-visualization}.d shows a carotid artery with a thickened wall for large parts of the CCA and ICA in the baseline examination. In the follow-up, the extracted plaque is localized the ICA. The unfolded plaque region displays a growth of the area with high VWT.

The last carotid artery (Fig.~\ref{fig:plaque-visualization}.e) shows a plaque that is restricted to the ICA. 

\subsection{Intensity Histograms for Plaque Composition Analysis}
\begin{figure}
    \centering
    \includegraphics[width=0.9\linewidth]{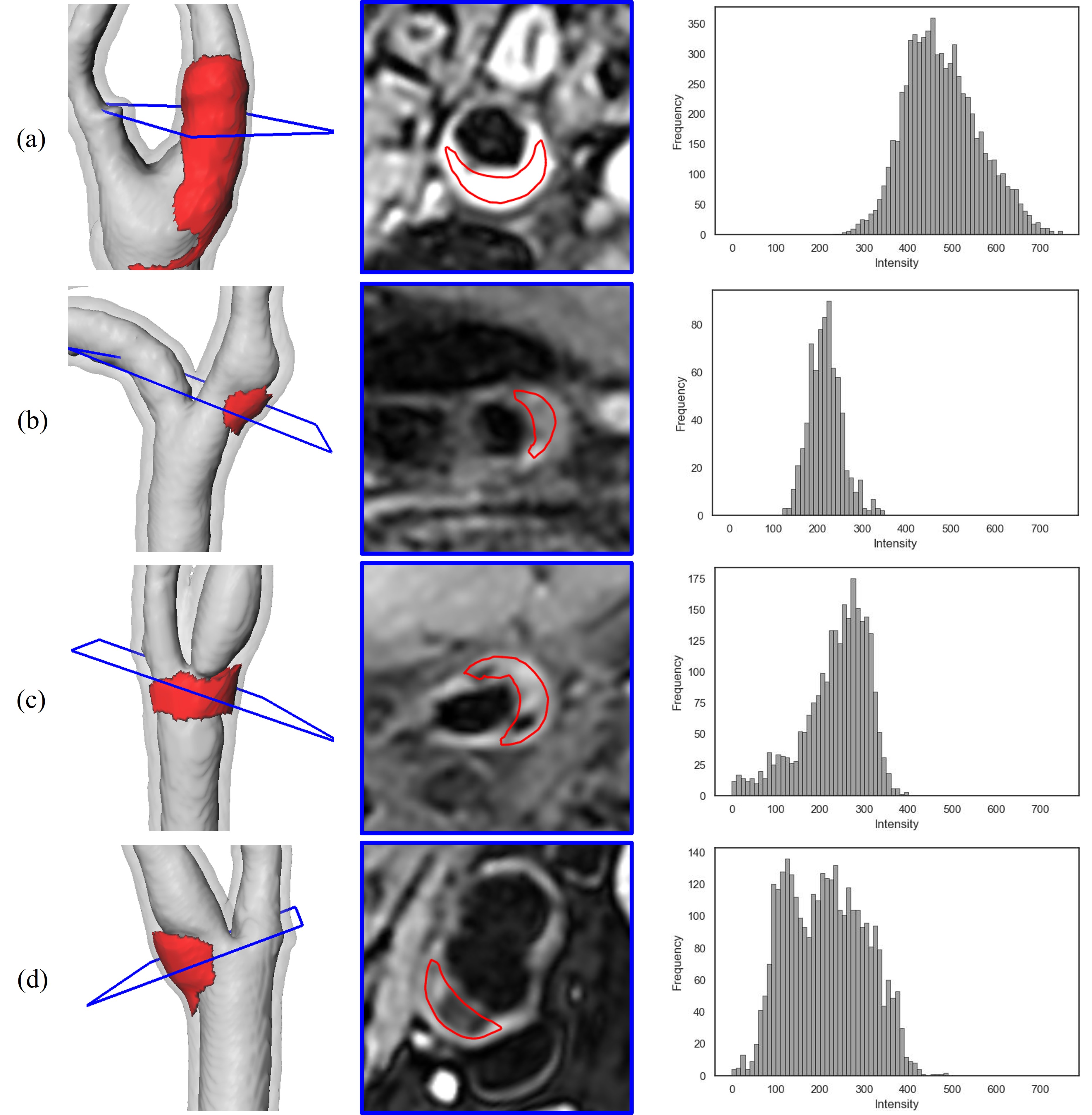}
    \caption{Analysis of T1-weighted MRI intensities within the plaque mesh. The plaque position (left), a 2D cross-section of the T1-weighted MRI (middle), and the histogram of the intensities of all voxels inside the plaque mesh.}
    \label{fig:intensity_histograms}
\end{figure}

Fig.~\ref{fig:intensity_histograms} shows the intensity distribution of the voxels inside the plaque mesh. The plaque in Fig.~\ref{fig:intensity_histograms}.a, includes hyperintense voxels resulting in a small histogram peak. The carotid artery wall of the plaque below (Fig.~\ref{fig:intensity_histograms}.b) shows two small hyperintensities, which are partly included in the plaque mesh. The corresponding histogram shows the hyperintensities as a small peak at intensity 325. Fig.~\ref{fig:intensity_histograms}.c shows a plaque with two calcifications represented as hypointense regions in the MRI images (see cross-section), and the histogram shows a high skewness. Fig.~\ref{fig:intensity_histograms}.d shows a carotid artery wall with a larger hypointense plaque region. Correspondingly, the histogram shows two peaks.

\section{Discussion}
The proposed method demonstrates reliable performance in the extraction of plaques in the ICA, CCA and bifurcation region. The mesh-based extraction approach has two advantages. It can extract and analyze a plaque with sub voxel resolution and it allows a direct quantification of plaque features using established mesh algorithms. However, the inner and outer wall mesh have to be created with the marching cubes algorithm and a smoothing, which can introduce errors. A wall segmentation method, that directly outputs a mesh, can mitigate this issue.

Using the case-specific threshold reduced the number of false positives in a healthy-set to zero, emphasizing the necessity to consider differences between subjects for an optimal solution. The computation of the case-specific threshold requires to choose a reasonable $k$. As the extracted plaque showed only small differences for $k\in\{2.5,3,3.5\}$ we consider the method to be robust to small changes of $k$. An interesting finding was that for $k\in\{4, 4.5\}$ plaques with a large volume are no longer extracted while plaques with a smaller volume were still extracted. This happens, because the initialization of $X_{normal}$ (Eq.~\ref{eq:iterative_threshold_exp2}) leads to a high deviation of VWT and a high initial plaque threshold if large parts of the vessel wall are thickened. This needs to be considered if this method is applied on 3D vessel wall segmentation where a large portion of the vessel wall is covered with plaque. In such cases, a global threshold can be preferable.

\paragraph{Quantitative and Visual Assessment of Plaque Properties}
Our method enables a comprehensive quantitative and visual assessment of plaque properties in 3D, providing valuable insights into atherosclerosis progression.

Previous studies have measured plaque volume by manually or semi-automatically segmenting the carotid wall in each MRI slice and used the vessel wall volume \cite{brunner2021associations, van2022carotid}. However, this method measures the wall volume instead of the plaque volume and requires manual segmentation. Our approach enables the automatic quantification of geometric features such as plaque volume, plaque area, and compactness. 
The relevance of plaque volume for stroke risk assessment has been shown before \cite{saba2021review, brunner2021associations, van2022carotid}, but further studies are needed to determine the influence of the plaque area and compactness. We observed that most plaques exhibit small volumes, typically less than 50~$mm^3$, but the method was capable of extracting plaques with higher volumes up to 847.750~$mm^3$.

Surface unfolding is a common technique in medical image analysis \cite{kreiser2018survey}, and Eulzer et al. \cite{eulzer2021automatic} successfully applied it to the surface of the carotid artery. Our approach improves this by focusing exclusively on the relevant plaque region. This offers a clearer understanding of plaque morphology, proving effective across various plaque shapes. However, comparing different subjects remains challenging.

Lastly, we analyzed the intensity distribution of voxels within the 3D plaque mesh. Although T1-weighted MRI intensities are not direct quantitative tissue measurements, patterns of hyperintensities and hypointensities can be seen in intensity histograms. These histograms, representing the complete plaque, are independent of cross-sectional placement and provide an overview of the intensity distribution.

\paragraph{Applicability} The method extracts plaques of varying sizes, at different positions and of different shapes.
The presented processing pipeline includes separate modules for segmentation and quantitative analysis which can be adapted to enable plaque extraction and analysis in other tomographic imaging modalities and vessel systems. If a 3D vessel wall segmentation exists, only $k$, which is used for the computation of the plaque threshold, has to be adapted. 

\paragraph{Conclusion and Future Work} In conclusion, this method provides reliable and specific plaque quantification.  Enhancements in vessel wall segmentation can be integrated into the existing pipeline to improve method accuracy. Further investigation is required to assess the clinical utility of the quantitative parameters for atherosclerosis progression monitoring and risk prediction. Expanding the method's application to other imaging modalities and vascular structures could broaden its versatility and clinical relevance.

\paragraph{Limitations}
The method is only based on the distance encoding and cannot extract a plaque if the thickness of the vessel wall is not increased.
The plaque mesh is extracted from the 3D vessel wall segmentation. Therefore, it is strongly influenced by segmentation quality.

\section*{Compliance with Ethical Standards}
\paragraph{Funding:}
This work is funded by the German Research Foundation (GRK2260, BIOQIC)\newline
Prof. Dr. Andreas Harloff and Dr. Christoph Strecker have received funding from German Research Foundation grant \#HA5399/5-1. \newline
Prof. Dr. Andreas Harloff is supported by the Berta-Ottenstein-Program for Advanced Clinician Scientists, Faculty of Medicine, University of Freiburg, Germany.\newline
Prof. Dr.-Ing. Anja Hennemuth received funding from German Research Foundation grant \#HE7312/7-1. \newline

\paragraph{Conflict of Interest: }
The authors have no relevant financial or non-financial interests to disclose.

\paragraph{Informed Consent:}
The data acquisition study \cite{strecker2020carotid} received approval from the Ethics Committee of the University of Freiburg (Approval No. 463/13). Written informed consent was obtained from all participants.

\bibliography{sn-bibliography}

\end{document}